\documentclass[11pt]{article}

\usepackage[final]{acl}

\usepackage{times}
\usepackage{latexsym}

\usepackage[T1]{fontenc}

\usepackage[utf8]{inputenc}

\usepackage{microtype}

\usepackage{inconsolata}

\usepackage{hyperref}
\usepackage{subcaption}
\usepackage{graphicx}
\usepackage{makecell}
\usepackage{multicol}
\usepackage{multirow}
\usepackage{threeparttable}
\usepackage{arydshln}
\usepackage{array}
\usepackage{algorithm}
\usepackage{algpseudocode}
\newcolumntype{C}[1]{>{\centering\arraybackslash}p{#1}}

\title{FBK's Long-form SpeechLLMs for IWSLT 2026 Instruction Following}

\author{
 \textbf{Zhihang Xie\textsuperscript{1,2}},
 \textbf{Marco Gaido\textsuperscript{1}},
 \textbf{Sara Papi\textsuperscript{1}},
 \textbf{Matteo Negri\textsuperscript{1}},
 \textbf{Luisa Bentivogli\textsuperscript{1}}
\\
\\
 \textsuperscript{1}Fondazione Bruno Kessler,
 \textsuperscript{2}University of Trento
}

\begin{document}
\maketitle
\begin{abstract}
This paper describes our submission to the IWSLT 2026 Instruction Following
shared task.
SpeechLLMs are developed for both short-form and long-form speech instruction
following under constrained settings.
For the short track, strong performance is achieved on MCIF, with a SIFS score
of 2.0708.
For the long track, three speech segmentation methods are explored, and
the HIFS score is introduced to account for unstable long-form generation.
Experimental results show that fixed 30-second segmentation provides the most
robust long-form performance, achieving the highest HIFS score of 2.0663.
Further analysis shows that hallucination mainly manifests as repetitive
insertions in generated outputs, substantially affecting ASR and SSUM, while
short-form capabilities are largely retained after long-form extension.
\end{abstract}

\section{Introduction}

Speech large language models (SpeechLLMs) extend the instruction-following (IF)
capabilities of large language models to spoken inputs, where a speech encoder
is connected to an LLM decoder through a modality
adapter~\cite{chen_llast_2024, huang_investigating_2024}.
This architecture supports multiple speech-to-text tasks with natural-language
instructions, including automatic speech recognition (ASR), speech translation
(ST), spoken question answering (SQA), and speech summarization (SSUM).
Compared with task-specific systems, instruction-following
SpeechLLMs~\cite{fathullah_audiochatllama_2024, lee_naver_2025} provide a
flexible interface for multilingual and multi-task speech processing, while
allowing spoken inputs to benefit from the generation abilities of pretrained
LLMs.

The International Conference on Spoken Language Translation (IWSLT) 2026
Instruction Following shared
task\footnote{\href{https://iwslt.org/2026/instruction-following}{IWSLT 2026
Instruction Following page}} evaluates speech-based instruction-following
systems in short-form and long-form settings.
The short track focuses on short-form speech and covers ASR, ST, and SQA
across languages.
The long track extends this setting to longer audio, requiring systems to
perform ASR, ST, SQA, SSUM, and audio chaptering (ACHAP).
Together, the two tracks provide a benchmark for assessing instruction-following
ability and robustness to long-form speech.

Despite recent progress in SpeechLLMs, long-form speech processing remains
challenging.
Long-form inputs introduce high computational cost, long acoustic token
sequences, discourse-level context modeling, and hallucination risks.
In IWSLT 2025, KIT~\cite{koneru_kits_2025} was the only
participant in the long track~\cite{abdulmumin_findings_2025}.
For SpeechLLMs, longer speech sequences place greater pressure on the LLM
context and may lead to unstable generation.
It therefore remains unclear how effectively short-form SpeechLLMs can be
extended to long-form settings, how much short-form ability is retained, and
which segmentation strategies are most effective for long-form inference.

Under the constrained setting, this study examines three hypotheses:
(1) long-form extension can improve long-form performance while preserving
competitive short-form capability;
(2) speech segmentation substantially affects
long-form instruction following;
and (3) hallucination-aware evaluation is
needed for reliable comparison of long-form systems.
The main contributions are:

\begin{figure*}[t]
  \centering
  \begin{subfigure}{0.325\textwidth}
    \centering
    \includegraphics[width=1.08\linewidth]{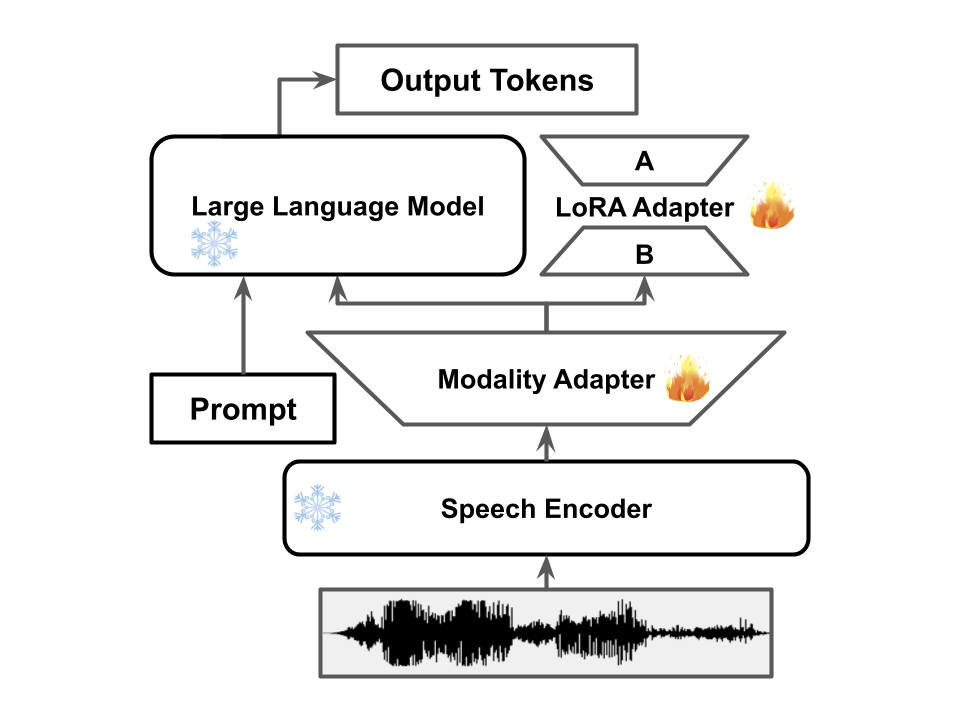}
    \caption{Model architecture}
    \label{fig:model-architecture}
  \end{subfigure}
  \hfill
    \begin{subfigure}{0.325\textwidth}
    \centering
    \includegraphics[width=1.08\linewidth]{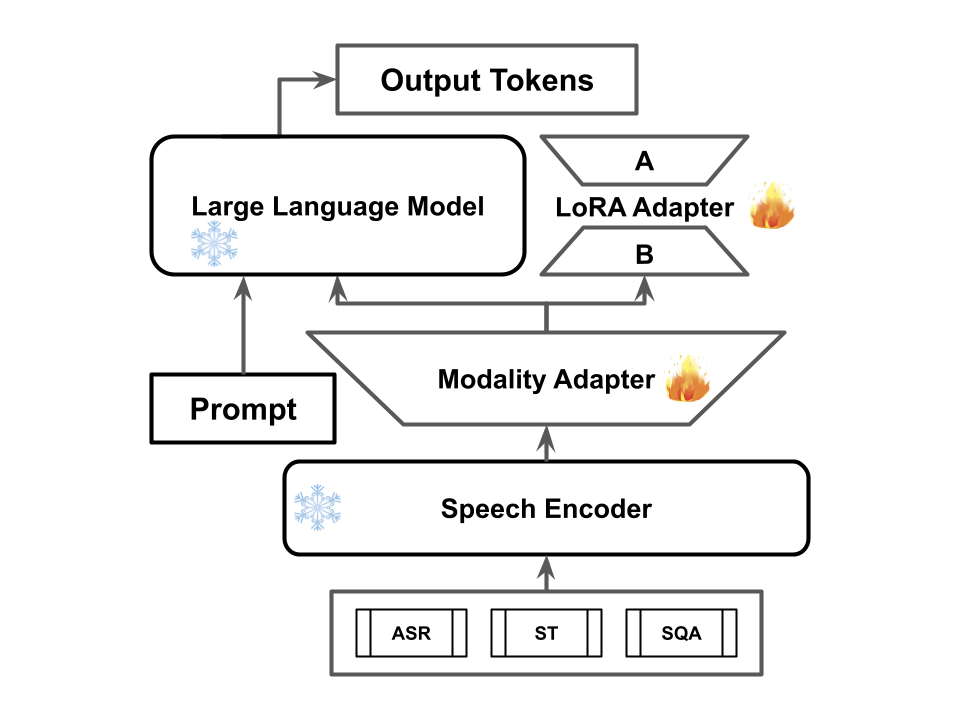}
    \caption{Short-form fine-tuning}
    \label{fig:short-form-finetuning}
  \end{subfigure}
  \hfill
  \begin{subfigure}{0.325\textwidth}
    \centering
    \includegraphics[width=1.08\linewidth]{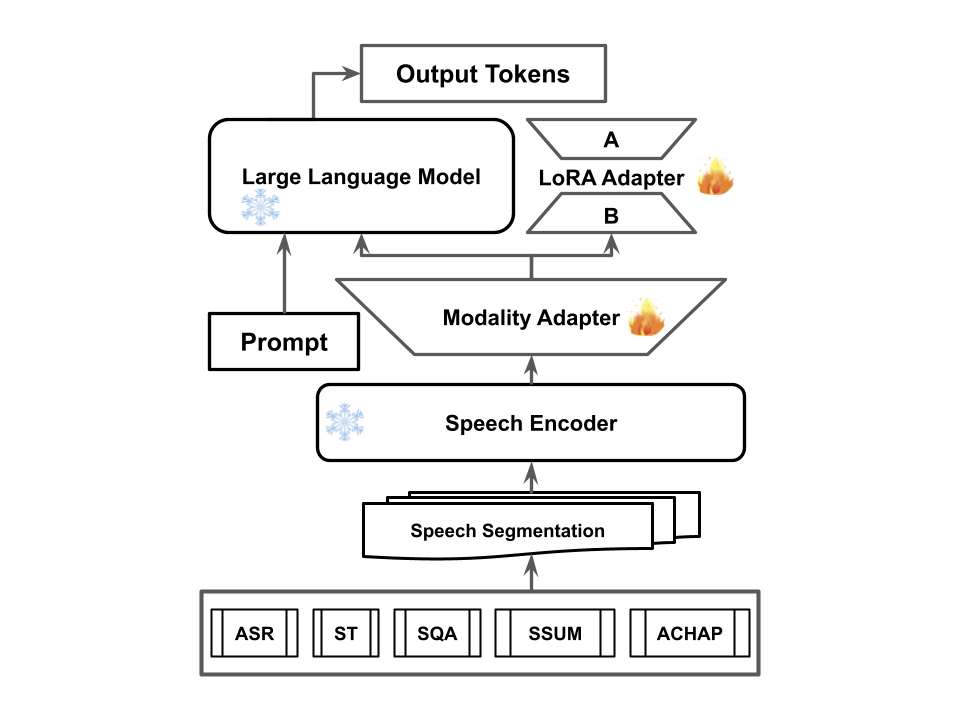}
    \caption{Long-form extension}
    \label{fig:long-form-extension}
  \end{subfigure}
  \caption{Model architecture and training pipeline of the SpeechLLMs for the
    short and long tracks.}
\end{figure*}

\begin{itemize}
\item Practical and effective SpeechLLM solutions for short-form and long-form
instruction following under constrained settings.
\item Empirical comparisons across three speech segmentation strategies:
fixed-window, CRDNN-based, and hybrid segmentation.
\item Detailed evaluation analysis of hallucination effects and short-form
capability retention after long-form extension.
\end{itemize}

\section{Model Architecture}

In the constrained setting, SpeechLLMs are required to be built upon two
pretrained models,
\textsc{SeamlessM4T-v2-large}\footnote{\href{https://huggingface.co/facebook/seamless-m4t-v2-large}{SeamlessM4T-v2-large
    on Huggingface}} and
\textsc{Qwen3-4B-Instruct}\footnote{\href{https://huggingface.co/Qwen/Qwen3-4B-Instruct-2507}{Qwen3-4B-Instructon
    Huggingface}}.
As illustrated in Figure~\ref{fig:model-architecture}, our SpeechLLMs comprise a
speech encoder, a modality adapter, and an LLM decoder.
The model architectures are identical for both the short track and long track,
where the finetuned short-form SpeechLLM can be extended to long-form directly.

\subsection{Speech Encoder}

The speech encoder is adopted from \textsc{SeamlessM4T-v2-large} and serves as
the acoustic front end of the model, transforming raw speech into intermediate
speech representations for downstream processing~\cite{seamlessm4t_2023}.
The feature extractor operates on 16 kHz audio and converts raw waveforms into
80-dimensional log-Mel filterbank features.
With a 10 ms hop size, these features are extracted at 100 Hz.
A stride of 2 is then applied in the frontend, such that two consecutive frames
are combined into a 160-dimensional representation before feature projection,
reducing the temporal resolution to 50 Hz.
These acoustic features are provided to the model together with an attention
mask indicating valid and padded positions.
In this way, the feature extractor forms the interface between speech input and
the encoder stack.

The speech encoder is built upon the extracted acoustic features and comprises
a feature projection module followed by a Conformer
stack~\cite{koluguri_investigating_2023}.
The feature projection maps the 160-dimensional frontend features to the model
hidden size of 1024, and the Conformer encoder then processes them into
high-level speech representations while preserving the temporal resolution of
50 Hz.
In the released large configuration, the encoder comprises 24 Conformer layers,
each with 16 attention heads and a feed-forward network of intermediate
dimension 4096.
Relative positional representations~\cite{dai_transformer-xl_2019} and a
depthwise convolution with kernel size 31 are incorporated in each Conformer
block, enabling the encoder to capture both long-range contextual dependencies
and local acoustic structure effectively.

\subsection{Modality Adapter}

The modality adapter comprises the adapter inherited from
\textsc{SeamlessM4T-v2-large} for temporal compression and a linear adapter for
projection into the hidden space of the LLM decoder.

Given 50 Hz speech representations from the speech encoder, an intermediate
feed-forward module first transforms the encoder outputs at the 1024-dimensional
hidden size before passing them to the inherited adapter for temporal
compression.
In this adapter, the residual and self-attention branches each apply a
one-dimensional convolution with kernel size 8, stride 8, and padding 4,
followed by a gated linear unit.
This design reduces the sequence length by approximately a factor of 8, yielding
an effective frame rate of about 6.25 Hz while preserving the 1024-dimensional
hidden size.

The adapter layer incorporates self-attention and a feed-forward network with
an intermediate dimension of 4096, allowing the compressed representations to be
refined after subsampling.
Following the inherited adapter, a linear adapter is introduced to project the
compressed speech representations from 1024 to 2560.
This projection aligns the speech representations with the decoder input space,
so that the resulting speech tokens have the same dimensionality as the text
tokens.

\subsection{LLM Decoder}

The LLM decoder is inherited from \textsc{Qwen3-4B-Instruct} and leverages its
strong instruction-following ability for general reasoning and token
generation.
To incorporate speech modality, the prepending fusion mechanism
\cite{lam_prepending_2025} is employed, whereby speech representations are
prepended to the text embeddings of the instruction.
In addition, parameter-efficient fine-tuning methods such as
LoRA~\cite{hu_lora_2021} are commonly used to adapt the LLM to downstream
tasks without updating the full set of model parameters
\cite{chen_llast_2024, microsoft_phi-4-mini_2025}.

\begin{figure*}[t]
  \centering
  \begin{subfigure}{0.4\textwidth}
    \centering
    \includegraphics[width=\linewidth]{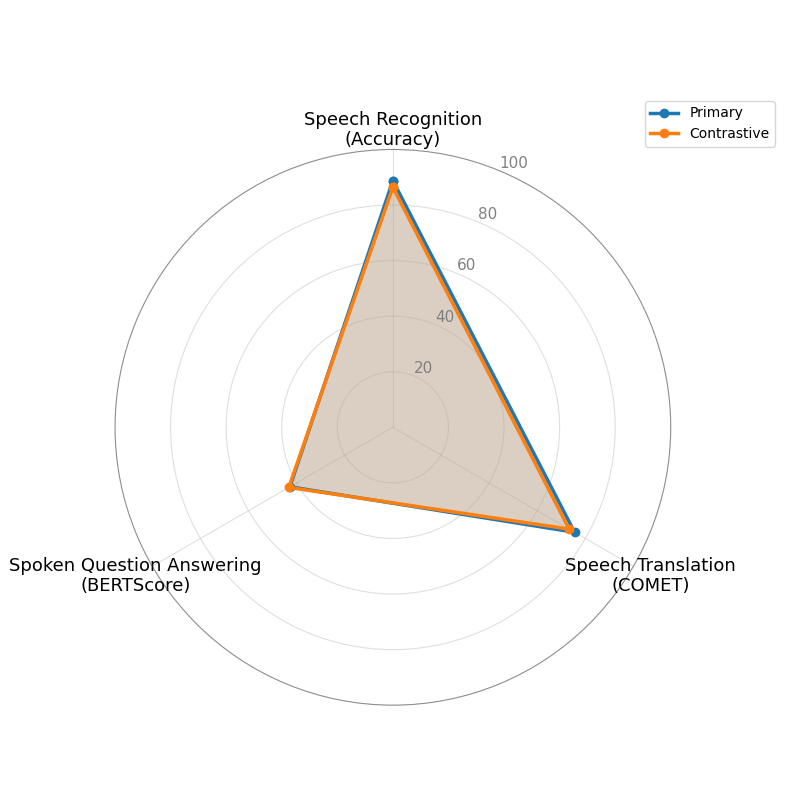}
    \caption{Short-form capabilities}
    \label{fig:short-form-capabilities}
  \end{subfigure}
  \hfill
    \begin{subfigure}{0.4\textwidth}
    \centering
    \includegraphics[width=\linewidth]{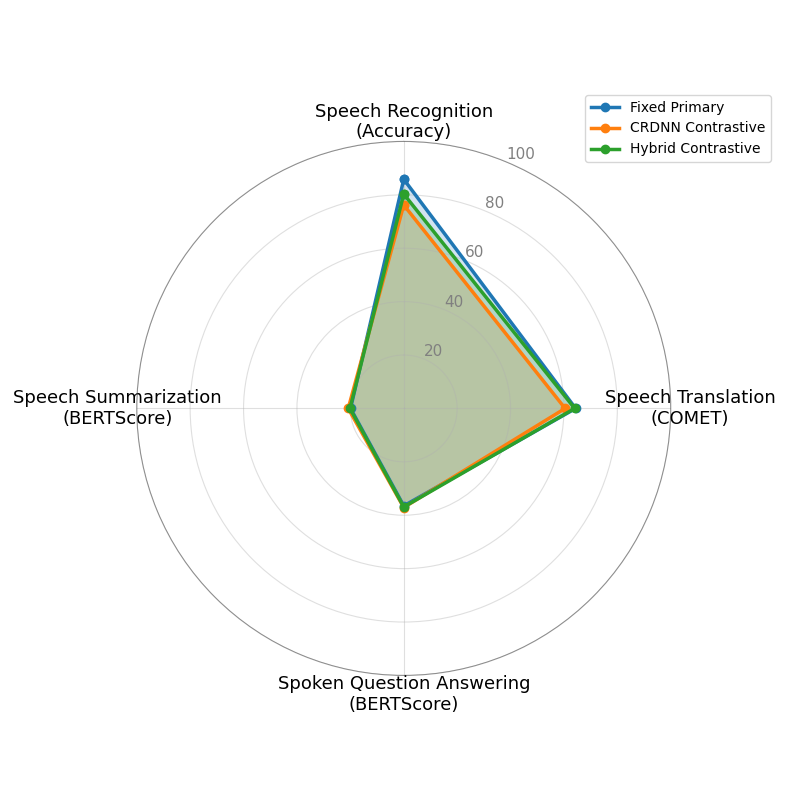}
    \caption{Long-form capabilities}
    \label{fig:long-form-capabilities}
  \end{subfigure}
  \caption{Radar charts of short-form and long-form capabilities, with ASR
    measured by accuracy, ST by mean COMET score, SQA and SSUM by BERTScore, and
    long-form scores penalized by hallucination rates.}
\end{figure*}

\section{Short Track}

\subsection{Corpora}

In the constrained setting, training is conducted using
CoVoST2~\cite{wang_covost_2020}, EuroParlST~\cite{sanchez_europarl_2020},
GigaST~\cite{ye_gigast_2023}, and LibriSQA~\cite{zhao_librisqa_2024}.
For data augmentation, synthetic data are generated with
\textsc{SeamlessM4T-v2-large} and \textsc{Qwen3-4B-Instruct}.
Validation is primarily performed on the MCIF~\cite{papi_mcif_2025} short-track
datasets.

The training corpora, summarized in Table~\ref{tab:short-corpora}, comprise
5,056,973 samples with a total duration of 8,436.63 hours and an average
duration of 6.01 seconds.
Because these corpora differ substantially in annotation format, language
coverage, and task supervision, dataset-specific processing is applied to
construct a unified multilingual instruction-following training set for ASR, ST,
and SQA.
Specifically, synthetic translations are generated with
\textsc{SeamlessM4T-v2-large} when target-language annotations are missing,
low-quality samples are filtered using COMET~\cite{rei_comet-22_2022}, and
synthetic question-answer pairs are generated with \textsc{Qwen3-4B-Instruct}
when question-answering supervision is unavailable or limited.
A small proportion of unanswerable examples is also introduced to improve
robustness in SQA.
The processing applied to each dataset is described below, and validation is
primarily performed on the MCIF~\cite{papi_mcif_2025} short-track datasets,
which provide a unified benchmark for multiple short-form tasks and languages.

\textbf{CoVoST2}\quad
The original dataset provides English transcriptions together with translations
from English into German and Chinese.
Since Italian translations are not included, they are generated from the English
transcriptions using \textsc{SeamlessM4T-v2-large}, after which low-quality
translations are removed using COMET~\cite{rei_comet-22_2022} scores below
0.85\footnote{Unbabel/wmt22-cometkiwi-da}.
To increase the amount of training data for spoken question answering, synthetic
question-answer pairs are generated with \textsc{Qwen3-4B-Instruct}.
In addition, to explicitly model unanswerable cases, five percent of the pairs
in each language are randomly sampled, their questions are replaced with random
questions drawn from the remaining samples, and their answers are set to ``Not
answerable’’ in the corresponding language.

\textbf{EuroParlST}\quad
The original dataset provides English transcriptions together with translations
from English into German and Italian.
Two training subsets are available, \texttt{train} and \texttt{train-noisy}, but
only the \texttt{train} subsets are used.
No synthetic translations from English to Chinese are generated.

\textbf{GigaST}\quad
The original dataset provides English transcriptions together with translations
from English into German and Chinese.
Five training subsets are available, \texttt{XS}, \texttt{S}, \texttt{M},
\texttt{L}, \texttt{XL}, but only the \texttt{M} subsets are used to maintain a
training data scale comparable to that of the other datasets.
Since the English transcriptions are provided entirely in uppercase, text
normalization is applied with \textsc{Qwen3-4B-Instruct} to convert them into
natural sentence form.
Low-quality translations are removed based on COMET scores lower than 0.85.

\textbf{LibriSQA}\quad
The original dataset provides English transcriptions together with
question-answer pairs.
To create synthetic translations, the English transcriptions are first
translated into German, Italian, and Chinese using
\textsc{SeamlessM4T-v2-large}, after which low-quality translations are removed
based on COMET thresholds of 0.8, 0.8, and 0.75, respectively.
Synthetic question-answer pairs are then generated using
\textsc{Qwen3-4B-Instruct}.
Following the same procedure as for CoVoST2, five percent of the pairs are
randomly selected, and their answers are replaced with ``Not answerable'' in the
corresponding language.

\subsection{Training Strategy}

The training pipeline is illustrated in Figure~\ref{fig:short-form-finetuning}.
The SpeechLLM contains 4.73B parameters in total and includes a newly introduced
two-layer linear module with an intermediate dimension of 3584.
Parameter-efficient adaptation is performed by training 112.2M parameters, with
LoRA applied to the query, key, and output projection modules using rank 8 and
alpha 16.
The model is optimized for two epochs with AdamW, using a total batch size of
128 and gradient clipping set to 1.0.
Separate learning rates are used for the base model and LoRA parameters, set to
$1e{-4}$ and $3e{-4}$, respectively, together with a cosine
learning-rate scheduler and a warmup phase covering 3\% of the total training
steps.
To improve robustness, data augmentation is applied, including speed
perturbation with factors [0.9, 1.0, 1.1] and SpecAugment with two time masks
and two frequency masks, where the maximum mask widths are 50 and 10,
respectively.
The model is trained on four NVIDIA A100 64GB GPUs for approximately two days,
The final checkpoint is used for evaluation and submission.

\subsection{System Evaluation}

\begin{table*}
  \centering
  \small
  \begin{tabular}{cccccc}
    \hline
    \textbf{Submission} & \textbf{Language}
    & \makecell{\textbf{ASR}\\\textbf{Accuracy}}
    & \makecell{\textbf{ST}\\\textbf{COMET}}
    & \makecell{\textbf{SQA}\\\textbf{BERTScore}}
    & \textbf{SIFS}\\
    \hline
    \multirow{5}{*}{\makecell{Primary\\Short-form}}
    & en-en & \textbf{0.8877} &               - & \textbf{0.4426} & - \\
    & en-de &               - & \textbf{0.7286} &          0.4152 & - \\
    & en-it &               - &          0.7496 &          0.4034 & - \\
    & en-zh &               - & \textbf{0.7869} & \textbf{0.4513} & - \\
    \cdashline{2-5}
    & score & \textbf{0.8877} & \textbf{0.7550} & 0.4281 & \textbf{2.0708} \\
    \hline
    \multirow{5}{*}{\makecell{Contrastive\\Long-form}}
    & en-en & 0.8640 &               - &          0.4413 & - \\
    & en-de &      - &          0.7036 & \textbf{0.4294} & - \\
    & en-it &      - & \textbf{0.7516} & \textbf{0.4128} & - \\
    & en-zh &      - &          0.7425 &          0.4466 & - \\
    \cdashline{2-5}
    & score & 0.8640 & 0.7326 & \textbf{0.4325} & 2.0291 \\
    \hline
  \end{tabular}
  \caption{Performance of the primary and contrastive systems on the MCIF
    short-form tasks.}
  \label{tab:mcif-short}
\end{table*}

Validation is performed on the MCIF~\cite{papi_mcif_2025} short-track datasets
covering multiple tasks and languages.
The Short-form Instruction-Following Score (SIFS) is computed by summing the
averaged task-level scores, as shown in Equation~\ref{eq:sifs}.
Here, $\mathcal{T}$ includes ASR, ST, and SQA; $\mathcal{L}_t$ denotes the
language pairs with valid results for task $t$; and $m_{\ell,t}$ denotes the
task-specific score, with ASR measured by $1-\mathrm{WER}$.

\begin{equation}
\label{eq:sifs}
\mathrm{SIFS}
=
\sum_{t \in \mathcal{T}}
\frac{1}{|\mathcal{L}_t|}
\sum_{\ell \in \mathcal{L}_t}
m_{\ell,t}.
\end{equation}

As shown in Table~\ref{tab:mcif-short} and
Figure~\ref{fig:short-form-capabilities}, the primary system achieves
competitive short-form performance, with a SIFS score of 2.0708, comprising an
ASR accuracy of 0.8877, an average ST COMET score of 0.7550, and an average SQA
BERTScore of 0.4281.

\subsection{System Submission}

As shown in Table~\ref{tab:mcif-short}, the primary short-form system achieves
the best overall performance, with a SIFS score of 2.0708.
The contrastive system is based on the long-form model and obtains a lower SIFS
score of 2.0291, but remains competitive on SQA.
This system is discussed further in Section~\ref{sec:long-eval}.

Table~\ref{tab:iwslt-short} shows the IWSLT 2026 short-form results
\cite{adelani-etal-2026-iwslt}.
The primary submission outperforms the contrastive submission on most tasks,
with comparable results on EN-EN and EN-IT.
Overall, the primary submission is more balanced across all tasks, whereas the
contrastive submission is less robust to unseen tasks.

\section{Long Track}

\subsection{Corpora}

In the constrained setting, the training corpora comprise LibriSQA for ASR, ST,
and SQA, Nutshell~\cite{zufle_nutshell_2025} for SSUM and
YTSeg~\cite{retkowski_text_2024} for ACHAP.
For data augmentation, \textsc{SeamlessM4T-v2-large} and
\textsc{Qwen3-4B-Instruct} are used to process the synthetic data.
Validation is performed on the MCIF~\cite{papi_mcif_2025} long-track datasets.

The training corpora, summarized in Table~\ref{tab:long-corpora}, comprise
825,256 samples with a total duration of 9,496.36 hours and an average
duration of 41.43 seconds.
Because Nutshell provides only abstracts without transcriptions, and YTSeg
contains highly noisy transcriptions, neither dataset provides reliable
supervision for long-form speech recognition and translation.
Initial zero-shot experiments further indicate that a SpeechLLM trained only on
short-form data generalizes poorly to long-form speech, showing substantial
performance degradation when required to process extended acoustic and
linguistic context.
This limitation is particularly important because speech recognition and
translation constitute the foundation for higher-level long-form speech
understanding tasks, such as speech summarization and audio chaptering.
To address this gap, long-form training data for speech recognition,
translation, and question answering are constructed from LibriSQA through
artificial concatenation~\cite{fox_updated_2024}.
The processing applied to each dataset is described below.
Validation is performed on the MCIF~\cite{papi_mcif_2025} long-track, which
provide a unified benchmark for the long-form tasks and languages.

\textbf{LibriSQA}\quad
The original dataset provides exclusively short-form samples, but these can be
reorganized into long-form examples by leveraging the chapter structure.
For artificial concatenation, utterance-level audio segments are first grouped
by chapter and then concatenated in sentence order, with a 0.5-second silence
inserted between consecutive utterances to preserve natural boundaries.
The transcriptions are processed in the same way:
Chinese translations are concatenated directly, whereas spaces are inserted
between consecutive sentences in English, German, and Italian to maintain
sentence boundaries.
Due to low-quality translations are removed during preprocessing, the
resulting concatenated chapters in German, Italian, and Chinese are sometimes
incomplete and therefore do not fully correspond to their English
counterparts.
Since question-answer pairs are provided at the utterance level, only one pair
is randomly selected for each concatenated sample; otherwise, the total
training duration would exceed 10,000 hours and become computationally
prohibitive.
The maximum duration at chapter level is limited to 10 minutes to reduce
computational cost.

\textbf{Nutshell}\quad
The original dataset provides English abstracts from scientific talks, and
translations into German, Italian, and Chinese are generated using
\textsc{SeamlessM4T-v2-large}.
Low-quality translations at abstract level are removed using COMET scores
below 0.75, 0.75, and 0.8, respectively.
The maximum duration at abstract level is limited to 15 minutes to reduce
computational cost.

\textbf{YTSeg}\quad
The original dataset provides English transcriptions and chapter annotations
for YouTube videos.
Chapter titles and timestamps are automatically generated using
\textsc{Whisper} and then merged into multiple target formats.
The chapter titles in the \texttt{target\_text\_ref} field are extracted and
processed to insert newlines and Markdown formatting.
Because the videos often contain noisy acoustic content, such as music and
background noise, VAD is applied to estimate voiced speech duration, and
samples with speech ratios below 0.7 are removed.
Translations into German, Italian, and Chinese are generated using
\textsc{SeamlessM4T-v2-large}.
No COMET-based filtering is applied to preserve complete chapter information.
The maximum duration at chapter level is limited to 10 minutes, shorter than
that used for Nutshell because chaptering requires substantially longer target
texts.

\begin{table*}
  \centering
  \small
  \begin{tabular}{ccccccc}
    \hline
    \textbf{Submission} & \textbf{HAL} & \textbf{WER} & \textbf{HIT} & \textbf{SUB} & \textbf{INS} & \textbf{DEL} \\
    \hline
    Fixed 30s & \textbf{+/-} & 0.1418 & 14,792 & 1,124 & 376 & 881 \\
    \hline
    \multirow{2}{*}{CRDNN}
    & \textbf{-} & 0.1579 & 13,037 &   983 &    271 & 1,139 \\
    & \textbf{+} & 1.0158 & 13,703 & 1,938 & 13,969 & 1,156 \\
    \hline
    \multirow{2}{*}{Hybrid}
    & \textbf{-} & 0.1569 & 13,828 & 1,054 &    312 & 1,149 \\
    & \textbf{+} & 0.4341 & 14,346 & 1,290 &  4,841 & 1,161 \\
    \hline
  \end{tabular}
  \caption{ASR WER breakdown for different segmentation strategies with and
    without hallucination (HAL).}
  \label{tab:mcif-wer}
\end{table*}

\begin{table*}
  \centering
  \scriptsize
  \begin{tabular}{ccccccccccc}
    \hline
    \multirow{2}{*}{\textbf{Submission}}
    & \multirow{2}{*}{\textbf{Language}}
    & \multicolumn{2}{c}{\textbf{ASR}}
    & \multicolumn{2}{c}{\textbf{ST}}
    & \multicolumn{2}{c}{\textbf{SQA}}
    & \multicolumn{2}{c}{\textbf{SUM}} \\
    & & \textbf{Accuracy} & \textbf{HAL}
    & \textbf{COMET} & \textbf{HAL}
    & \textbf{BERTScore} & \textbf{HAL}
    & \textbf{BERTScore} & \textbf{HAL}
    & \textbf{HIFS} \\
    \hline
    \multirow{5}{*}{\makecell{Primary\\Fixed 30s}}
    & en-en & 0.8582 & 0/21 &      - &    - & 0.3755 & 1/220 & 0.1850 &  7/100 & - \\
    & en-de &      - &    - & 0.7193 & 3/21 & 0.3699 & 0/220 & 0.2008 & 25/100 & - \\
    & en-it &      - &    - & 0.7546 & 1/21 & 0.3356 & 4/220 & 0.2203 & 31/100 & - \\
    & en-zh &      - &    - & 0.7365 & 4/21 & 0.3937 & 4/220 & 0.3586 & 10/100 & - \\
    \cdashline{2-11}
    & score & \textbf{0.8582} & \textbf{0/21} & \textbf{0.6438} & \textbf{8/63} & 0.3649 & 9/880 & 0.1993 & 73/400 & \textbf{2.0663} \\
    \hline
    \multirow{5}{*}{\makecell{Contrastive\\CRDNN}}
    & en-en & 0.8421 & 2/21 &      - &    - & 0.3755 & 0/220 & 0.1932 & 11/100 & - \\
    & en-de &      - &    - & 0.7035 & 2/21 & 0.3795 & 0/220 & 0.1913 & 21/100 & - \\
    & en-it &      - &    - & 0.7749 & 5/21 & 0.3320 & 2/220 & 0.2205 & 28/100 & - \\
    & en-zh &      - &    - & 0.7696 & 5/21 & 0.4041 & 2/220 & 0.3573 &  2/100 & - \\
    \cdashline{2-11}
    & score & 0.7619 & 2/21 & 0.6044 & 12/63 & \textbf{0.3711} & \textbf{4/880} & \textbf{0.2080} & \textbf{62/400} & 1.9454 \\
    \hline
    \multirow{5}{*}{\makecell{Contrastive\\Hybrid}}
    & en-en & 0.8431 & 1/21 &      - &    - & 0.3728 & 0/220 & 0.1975 & 11/100 & - \\
    & en-de &      - &    - & 0.7043 & 1/21 & 0.3747 & 0/220 & 0.2024 & 28/100 & - \\
    & en-it &      - &    - & 0.7761 & 4/21 & 0.3333 & 3/220 & 0.2335 & 38/100 & - \\
    & en-zh &      - &    - & 0.7678 & 4/21 & 0.4038 & 2/220 & 0.3545 &  4/100 & - \\
    \cdashline{2-11}
    & score & 0.8030 & 1/21 & 0.6402 & 9/63 & 0.3691 & 5/880 & 0.2016 & 81/400 & 2.0139 \\
    \hline
  \end{tabular}
  \caption{Performance of different speech segmentation strategies on the MCIF
    long-form tasks, together with hallucination (HAL) statistics, where the
    fixed 30-second window strategy achieves the best overall performance.}
  \label{tab:mcif-strategy}
\end{table*}

\subsection{Training Strategy}

The training pipeline is illustrated in Figure~\ref{fig:long-form-extension}.
The model follows the same architecture as the short-track system and is
optimized using the same hyperparameters.
Speed perturbation is disabled, while SpecAugment is kept unchanged.
The gradient accumulation step is set to 8, and the fixed batch-size strategy is
replaced with a dynamic bucket sampler to better accommodate variable-length
speech inputs.
For long-form audio, speech segmentation is applied before training.
The model is trained on four NVIDIA A100 64GB GPUs for approximately three days.
Due to time constraints, only one of the two scheduled training epochs is
completed by the submission deadline.
The final checkpoint is used for evaluation and submission.

\subsubsection{Dynamic Bucket Sampler}

To improve training efficiency under large variation in audio duration, we
employ a dynamic bucket-based distributed batch sampler.
The sampler assigns each sample to a duration bucket defined by a set of
boundaries, and each bucket is associated with a dedicated per-replica batch
size.
This allows shorter utterances to be grouped into larger batches and longer
utterances into smaller ones, reducing padding overhead and improving hardware
utilization.
At the beginning of each epoch, sample indices are deterministically shuffled
and grouped by bucket.
Within each bucket, synchronized global chunks are formed and evenly partitioned
across replicas so that all workers process samples from the same duration range
at each step.

When the number of samples in a bucket is not divisible by the global chunk
size, the sampler either drops the remainder or pads the bucket by
deterministically resampling indices from the same bucket during training.
It further supports both sequential and round-robin scheduling across buckets,
enabling either contiguous or interleaved traversal of different duration ranges.
The sampler maintains explicit state, including the current epoch and iterator position,
making it fully compatible with resumable and stateful distributed training.

\subsubsection{Speech Segmentation}

\begin{algorithm}[t]
  \caption{\textsc{DivideAndConquer}}
  \label{algorithm}
  \small
  \begin{algorithmic}[1]
    \Require Pause list $\mathcal{P}$, left boundary $L$, right boundary $R$, maximum duration $D$
    \Ensure Refined segment list within $[L, R]$

    \Function{DivideAndConquer}{$\mathcal{P}, L, R, D$}
    \If{$R \leq L$}
    \State \Return $[\ ]$
    \EndIf
    \If{$R - L \leq D$}
    \State \Return $[(L, R)]$
    \EndIf
    \State $p^\star \gets$ \Call{FindLongestPause}{$\mathcal{P}, L, R$}
    \If{$p^\star = \texttt{None}$}
    \State \Return $[(L, R)]$
    \EndIf
    \State $(p_s, p_e) \gets p^\star$
    \State $\mathcal{R} \gets [\ ]$
    \If{$L < p_s$}
    \State $\mathcal{R} \gets \mathcal{R} \cup$ \Call{DivideAndConquer}{$\mathcal{P}, L, p_s, D$}
    \EndIf
    \If{$p_e < R$}
    \State $\mathcal{R} \gets \mathcal{R} \cup$ \Call{DivideAndConquer}{$\mathcal{P}, p_e, R, D$}
    \EndIf
    \State \Return $\mathcal{R}$
    \EndFunction
  \end{algorithmic}
\end{algorithm}

To process long-form speech with a model trained primarily on shorter utterances,
three segmentation strategies are explored.
The first strategy uses fixed windows, where each audio is divided into
consecutive non-overlapping segments with durations of 15, 30, 45, or 60
seconds.
This setting provides a simple and controlled baseline for analyzing the effect
of segment length on downstream performance.
Shorter windows can reduce memory and context burden within each segment, while
longer windows preserve more local context but increase the risk of deletion and
truncation errors.

The second strategy applies voice activity detection using the SpeechBrain CRDNN
model to produce speech-based segments~\cite{ravanelli_speechbrain_2021}.
The \texttt{activation\_th}, \texttt{deactivation\_th}, and \texttt{speech\_th}
thresholds control speech onset detection, speech offset detection, and speech
region selection, respectively.
The default setting, (0.5, 0.25, 0.5), is used as the baseline; however, it may
produce fragmented segments and limit the long-form context.
Therefore, less aggressive variants are evaluated by relaxing these thresholds
to (0.3, 0.2, 0.3) and (0.2, 0.1, 0.2), encouraging the detector to preserve
more continuous speech regions and produce longer segments with broader context.

The third strategy adopts a hybrid segmentation approach inspired by
\cite{gaido_beyond_2021, potapczyk_srpols_2020}, combining pause-based
segmentation with a duration constraint.
The double-check step in the original pipeline is replaced with a
divide-and-conquer algorithm, as shown in Algorithm~\ref{algorithm}.
For segments exceeding the maximum allowed duration, the algorithm identifies
the longest pause within the current interval and recursively splits the segment
at that point.
If no valid pause is found, the segment is kept unchanged.

\subsection{System Evaluation}
\label{sec:long-eval}

The long-form system is evaluated from four perspectives:
short-form capability retention, hallucination, the evaluation metric, and
speech segmentation.
The analysis is conducted on the MCIF long-form tasks, covering ASR, ST, SQA,
and SSUM, with task-specific scores and hallucination statistics reported
for a comprehensive assessment.

\subsubsection{Short-form}

The long-form model is also evaluated on the MCIF short-form tasks to examine
whether long-form extension preserves short-form capabilities.
As shown in Table~\ref{tab:mcif-short}, the resulting contrastive system
achieves a SIFS score of 2.0291, lower than the primary short-form score of
2.0708 but still competitive.
This degradation may be partially explained by the training setup: to reduce
computational cost, only LibriSQA short-form data are used instead of the full
short-form training set.
Overall, these results suggest that short-form capabilities can largely be
retained after long-form extension.

\subsubsection{Hallucination}

Hallucination has a substantial impact on long-form evaluation, particularly for
ASR and SSUM, and can significantly distort the apparent effectiveness of a
segmentation strategy.
As shown in Table~\ref{tab:mcif-hallucination}, under CRDNN segmentation,
including hallucinated outputs increases ASR WER from 0.1580 to 1.0161 for
English, while also reducing ST COMET and SSUM BERTScore across target
languages.
In contrast, SQA is less affected, with only small differences in BERTScore
after hallucinated outputs are included.
This suggests that hallucination introduces task-dependent degradation, with
generation-heavy tasks being vulnerable.

The ASR error breakdown in Table~\ref{tab:mcif-wer} further shows that
hallucination mainly increases insertion errors.
For CRDNN segmentation, the number of insertions rises sharply from 271 to
13,969 when hallucinated outputs are included, leading to a much higher WER.
A similar but smaller trend is observed for the hybrid strategy, where
insertions increase from 312 to 4,841.
Manual inspection of the model outputs shows substantial hallucinated
repetitions, which explains why excessive insertions become the dominant source
of degradation.
Inspired by the post-editing strategy in prior work~\cite{koneru_kits_2025}, an
alternative regular-expression-based post-editing strategy is applied to remove
hallucinated repetitions and improve the reliability of long-form outputs before
evaluation.

\subsubsection{Evaluation Metric}

The Hallucination-Penalized Instruction-Following Score (HIFS) is computed by
applying a hallucination penalty to each available task score and then summing
the task-level averages, as shown in Equation~\ref{eq:hifs}.
Here, $\mathcal{T}$ denotes the set of evaluated tasks, including ASR, ST, SQA,
and SSUM; $\mathcal{L}t$ denotes the set of language pairs with valid results
for task $t$; $m{\ell,t}$ denotes the task-specific metric score; and
$h_{\ell,t}/n_{\ell,t}$ denotes the corresponding hallucination rate. For ASR,
$m_{\ell,t}$ is computed as $1-\mathrm{WER}$.

\vspace{-1em}

\begin{equation}
\label{eq:hifs}
\mathrm{HIFS}
=
\sum_{t \in \mathcal{T}}
\frac{1}{|\mathcal{L}_t|}
\sum_{\ell \in \mathcal{L}_t}
m_{\ell,t}
\left(1 - \frac{h_{\ell,t}}{n_{\ell,t}}\right).
\end{equation}

\subsubsection{Speech Segmentation}

Three speech segmentation strategies are compared for long-form inference:
fixed-window segmentation, CRDNN-based segmentation, and hybrid segmentation.
For fixed-window segmentation, Table~\ref{tab:mcif-window} shows that the
30-second window achieves the best overall performance, with the highest HIFS
score of 2.0663.
The best ASR accuracy and ST COMET score are also obtained with this setting,
while relatively low hallucination rates are maintained.
Although slightly better SQA and SSUM scores are achieved by the 15-second and
45-second windows, respectively, these task-specific gains do not lead to the
best overall performance.
This suggests that a 30-second window provides a better trade-off between
preserving sufficient context and avoiding unstable generation.
Adaptive segmentation is further evaluated using CRDNN-based VAD with different
parameter settings.
As shown in Table~\ref{tab:mcif-crdnn}, the $(0.5, 0.25, 0.5)$ setting achieves
the highest HIFS score of 1.9591, mainly due to its stronger ST performance.
In comparison, better SQA and SSUM scores and lower hallucination rates are
achieved by the $(0.3, 0.2, 0.3)$ setting, but its lower ST score results in a
slightly lower overall score.
These results indicate that CRDNN segmentation is sensitive to its parameter
settings, and that optimization for one task does not necessarily yield the best
overall long-form performance.
Finally, hybrid segmentation, which combines fixed-window and VAD-based
segmentation, is evaluated in Table~\ref{tab:mcif-hybrid}, and the scores are
visualized in Figure~\ref{fig:long-form-capabilities}.
The $(0.3, 0.2, 0.3)$ setting performs best among hybrid configurations,
reaching an HIFS score of 2.0139.
Compared with CRDNN segmentation, the overall score is improved, and the best
ASR accuracy and SQA BERTScore are obtained among hybrid settings.
However, this strategy remains below the fixed 30-second window in overall
performance.
These results suggest that adaptive segmentation can be beneficial but
requires careful tuning, while the fixed 30-second window provides the most
robust segmentation strategy for long-form inference.

\subsection{System Submission}

Based on the segmentation analysis, three systems are selected for submission:
fixed 30-second segmentation as the primary system, and CRDNN-based and hybrid
segmentation as contrastive systems.
As shown in Table~\ref{tab:mcif-strategy}, the primary system achieves the best
performance, with the highest HIFS score of 2.0663, as well as the best ASR
accuracy and ST COMET score.
Among the contrastive systems, CRDNN obtains the best SQA and SSUM scores, while
hybrid segmentation provides a stronger balance with an HIFS score of 2.0139.

Table~\ref{tab:iwslt-long} shows the IWSLT 2026 long-form results
\cite{adelani-etal-2026-iwslt}.
No segmentation method consistently performs best across all tasks and metrics.
Hybrid segmentation provides the most evident ASR improvement, whereas gains on
other tasks are less stable.
SSUM and ACHAP do not function reliably in this setting, with scores remaining
low or degenerate across segmentation methods.
These results highlight the difficulty of extending short-form to long-form
speech processing.

\section{Conclusion}

This paper presented SpeechLLM systems for short-form and long-form instruction
following.
Strong short-form performance was achieved, while long-form performance remained
sensitive to segmentation and hallucination.
Fixed 30-second segmentation emerged as the most robust strategy, while
hallucination mainly occurred as repetitive insertions that degraded ASR and
SSUM.
Long-form extension preserved much of the short-form capability, but better
segmentation and hallucination mitigation remain important future directions.

\section*{Limitations}

This work has several limitations.
First, only one epoch of long-form training is completed due to time and
computational constraints, which may limit adaptation to long-form speech.
Second, using only a subset of short-form data during long-form extension may
contribute to short-form degradation.
Third, only three SpeechBrain VAD parameters are explored, leaving broader
hyperparameter tuning and discourse-aware segmentation for future work.
Finally, regular-expression-based post-editing provides limited hallucination
mitigation, targeting surface-level repetitions but missing subtler
hallucinations or semantic inconsistencies.

\section*{Acknowledgments}

This paper has received funding from the European Union’s Horizon Europe programme under grant agreement No. 101213369 (project DVPS), and from the Digital Europe Programme (DIGITAL) under grant agreement No. 101198470 (project LLMs4EU). This work was also supported by the CINECA ISCRA-B programme through project LoST (Long-form Speech-to-Text Models via Sequence Compression), which provided access to high-performance computing resources and support.


\bibliography{main}

@inproceedings{huang_investigating_2024,
	title = {Investigating {Decoder}-only {Large} {Language} {Models} for {Speech}-to-text {Translation}},
  booktitle = {Proc. of Interspeech},
	author = {Huang, Chao-Wei and Lu, Hui and Gong, Hongyu and Inaguma, Hirofumi and Kulikov, Ilia and Mavlyutov, Ruslan and Popuri, Sravya},
	year = {2024},
}

@inproceedings{fathullah_audiochatllama_2024,
	title = {{AudioChatLlama}: {Towards} {General}-{Purpose} {Speech} {Abilities} for {LLMs}},
	booktitle = {Proc. of NAACL},
	author = {Fathullah, Yassir and Wu, Chunyang and Lakomkin, Egor and Li, Ke and Jia, Junteng and Shangguan, Yuan and Mahadeokar, Jay and Kalinli, Ozlem and Fuegen, Christian and Seltzer, Mike},
	year = {2024},
}

@inproceedings{abdulmumin_findings_2025,
	title = {Findings of the {IWSLT} 2025 {Evaluation} {Campaign}},
	booktitle = {Proc. of {IWSLT}},
	author = {Abdulmumin, Idris and others},
	year = {2025},
}

@inproceedings{seamlessm4t_2023,
	title = {{SeamlessM4T}: {Massively} {Multilingual} \& {Multimodal} {Machine} {Translation}},
	booktitle = {Proc. of WMT},
	author = {Barrault, Loïc and others},
	year = {2023},
}

@inproceedings{koluguri_investigating_2023,
	title = {Investigating {End}-to-{End} {ASR} {Architectures} for {Long} {Form} {Audio} {Transcription}},
  booktitle = {Proc. of Interspeech},
	author = {Koluguri, Nithin Rao and Kriman, Samuel and Zelenfroind, Georgy and Majumdar, Somshubra and Rekesh, Dima and Noroozi, Vahid and Balam, Jagadeesh and Ginsburg, Boris},
	year = {2023},
}

@inproceedings{dai_transformer-xl_2019,
	title = {Transformer-{XL}: {Attentive} {Language} {Models} beyond a {Fixed}-{Length} {Context}},
	booktitle = {Proc. of ACL},
	author = {Dai, Zihang and Yang, Zhilin and Yang, Yiming and Carbonell, Jaime and Le, Quoc and Salakhutdinov, Ruslan},
	year = {2019},
}

@inproceedings{lam_prepending_2025,
	title = {Prepending or {Cross}-{Attention} for {Speech}-to-{Text}? {An} {Empirical} {Comparison}},
  booktitle = {Proc. of NAACL},
	author = {Lam, Tsz Kin and Gaido, Marco and Papi, Sara and Bentivogli, Luisa and Haddow, Barry},
	year = {2025},
}

@inproceedings{hu_lora_2021,
	title = {{LoRA}: {Low}-{Rank} {Adaptation} of {Large} {Language} {Models}},
  booktitle = {Proc. of ICLR},
	author = {Hu, Edward J. and Shen, Yelong and Wallis, Phillip and Allen-Zhu, Zeyuan and Li, Yuanzhi and Wang, Shean and Wang, Lu and Chen, Weizhu},
	year = {2021},
}

@misc{chen_llast_2024,
	title = {{LLaST}: {Improved} {End}-to-end {Speech} {Translation} {System} {Leveraged} by {Large} {Language} {Models}},
	publisher = {arXiv},
	author = {Chen, Xi and Zhang, Songyang and Bai, Qibing and Chen, Kai and Nakamura, Satoshi},
	year = {2024},
}

@misc{microsoft_phi-4-mini_2025,
	title = {Phi-4-{Mini} {Technical} {Report}: {Compact} yet {Powerful} {Multimodal} {Language} {Models} via {Mixture}-of-{LoRAs}},
	publisher = {arXiv},
	author = {Microsoft},
	year = {2025},
}

@misc{wang_covost_2020,
	title = {{CoVoST} 2 and {Massively} {Multilingual} {Speech}-to-{Text} {Translation}},
	publisher = {arXiv},
	author = {Wang, Changhan and Wu, Anne and Pino, Juan},
	year = {2020},
}

@misc{ye_gigast_2023,
	title = {{GigaST}: {A} 10,000-hour {Pseudo} {Speech} {Translation} {Corpus}},
	publisher = {arXiv},
  author={Ye, Rong and Zhao, Chengqi and Ko, Tom and Meng, Chutong and Wang, Tao and Wang, Mingxuan and Cao, Jun},
	year = {2023},
}

@article{zhao_librisqa_2024,
  title={{LibriSQA}: A novel dataset and framework for spoken question answering with large language models},
  author={Zhao, Zihan and Jiang, Yiyang and Liu, Heyang and Wang, Yu and Wang, Yanfeng},
  journal={Trans. on Artificial Intelligence},
  year={2024},
}

@inproceedings{papi_mcif_2025,
	title = {{MCIF}: {Multimodal} {Crosslingual} {Instruction}-{Following} {Benchmark} from {Scientific} {Talks}},
  booktitle = {Proc. of ICLR},
	author = {Papi, Sara and Züfle, Maike and Gaido, Marco and Savoldi, Beatrice and Liu, Danni and Douros, Ioannis and Bentivogli, Luisa and Niehues, Jan},
	year = {2026},
}

@inproceedings{rei_comet-22_2022,
	title = {{COMET}-22: {Unbabel}-{IST} 2022 {Submission} for the {Metrics} {Shared} {Task}},
	booktitle = {Proc. of WMT},
	author = {Rei, Ricardo and C. de Souza, José G. and Alves, Duarte and Zerva, Chrysoula and Farinha, Ana C and Glushkova, Taisiya and Lavie, Alon and Coheur, Luisa and Martins, André F. T.},
	year = {2022},
}

@inproceedings{zufle_nutshell_2025,
	title = {{NUTSHELL}: {A} {Dataset} for {Abstract} {Generation} from {Scientific} {Talks}},
  booktitle = {Proc. of IWSLT},
	author = {Züfle, Maike and Papi, Sara and Savoldi, Beatrice and Gaido, Marco and Bentivogli, Luisa and Niehues, Jan},
	year = {2025},
}

@inproceedings{retkowski_text_2024,
	title = {From {Text} {Segmentation} to {Smart} {Chaptering}: {A} {Novel} {Benchmark} for {Structuring} {Video} {Transcriptions}},
	booktitle = {Proc. of ECACL},
	author = {Retkowski, Fabian and Waibel, Alexander},
	year = {2024},
}

@inproceedings{fox_updated_2024,
	title = {Updated {Corpora} and {Benchmarks} for {Long}-{Form} {Speech} {Recognition}},
	booktitle = {Proc. of {ICASSP}},
	author = {Fox, Jennifer Drexler and Raj, Desh and Delworth, Natalie and McNamara, Quinn and Miller, Corey and Jetté, Migüel},
	year = {2024},
}

@misc{ravanelli_speechbrain_2021,
	title = {{SpeechBrain}: {A} {General}-{Purpose} {Speech} {Toolkit}},
	publisher = {arXiv},
	author = {Ravanelli, Mirco and Parcollet, Titouan and Plantinga, Peter and Rouhe, Aku and Cornell, Samuele and Lugosch, Loren and Subakan, Cem and Dawalatabad, Nauman and Heba, Abdelwahab and Zhong, Jianyuan and Chou, Ju-Chieh and Yeh, Sung-Lin and Fu, Szu-Wei and Liao, Chien-Feng and Rastorgueva, Elena and Grondin, François and Aris, William and Na, Hwidong and Gao, Yan and Mori, Renato De and Bengio, Yoshua},
	year = {2021},
}

@inproceedings{gaido_beyond_2021,
	title = {Beyond {Voice} {Activity} {Detection}: {Hybrid} {Audio} {Segmentation} for {Direct} {Speech} {Translation}},
	booktitle = {Proc. of ICNLSP},
	author = {Gaido, Marco and Negri, Matteo and Cettolo, Mauro and Turchi, Marco},
	year = {2021},
}

@inproceedings{potapczyk_srpols_2020,
	title = {{SRPOL}'s {System} for the {IWSLT} 2020 {End}-to-{End} {Speech} {Translation} {Task}},
	booktitle = {Proc. of IWSLT},
	author = {Potapczyk, Tomasz and Przybysz, Pawel},
	year = {2020},
}

@inproceedings{koneru_kits_2025,
	title = {{KIT}'s {Offline} {Speech} {Translation} and {Instruction} {Following} {Submission} for {IWSLT} 2025},
  booktitle = {Proc. of IWSLT},
	author = {Koneru, Sai and Züfle, Maike and Nguyen, Thai-Binh and Akti, Seymanur and Niehues, Jan and Waibel, Alexander},
	year = {2025},
}

@inproceedings{lee_naver_2025,
	title = {{NAVER} {LABS} {Europe} {Submission} to the {Instruction}-following {Track}},
  booktitle = {Proc. of IWSLT},
	author = {Lee, Beomseok and Boito, Marcely Zanon and Besacier, Laurent and Calapodescu, Ioan},
	year = {2025},
}

@inproceedings{sanchez_europarl_2020,
	title = {Europarl-{ST}: {A} {Multilingual} {Corpus} for {Speech} {Translation} of {Parliamentary} {Debates}},
	booktitle = {Proc. of Interspeech},
	author = {Iranzo-Sánchez, Javier and Silvestre-Cerdà, Joan Albert and Jorge, Javier and Roselló, Nahuel and Giménez, Adrià and Sanchis, Albert and Civera, Jorge and Juan, Alfons},
	year = {2020},
}

@inproceedings{adelani-etal-2026-iwslt,
  title = {Speech Translation and Metrics in 2026: Findings of the IWSLT Campaign},
  author = {
  Adelani, David Ifeoluwa
                and Agostinelli, Victor
                and Anastasopoulos, Antonios
                and Bentivogli, Luisa
                and Bojar, Ond{\v{r}}ej
                and Brati{\`e}res, Sebastien
                and Carpuat, Marine
                and Cattoni, Roldano
                and Cettolo, Mauro
                and Chen, Lizhong
                and Federico, Marcello
                and Gaido, Marco
                and Gupta, Mahendra
                and Han, HyoJung
                and Hatami, Ali
                and Javorsk{\'y}, David
                and Jeon, Yejin
                and Kasztelnik, Marek
                and Laurent, Antoine
                and Liu, Danni
                and Luu, Nam
                and Ma, Min
                and Mach{\'a}{\v{c}}ek, Dominik
                and Maltais, Marie
                and Matusov, Evgeny
                and Maurya, Chandresh Kumar
                and McCrae, John P.
                and Meng, Chutong
                and Mohammad, Mohammadamini
                and Moslem, Yasmin
                and Murray, Kenton
                and Nakamura, Satoshi
                and Negri, Matteo
                and Niehues, Jan
                and Ojha, Atul Kr.
                and Ortega, John
                and Ouyang, Siqi
                and Papi, Sara
                and Pol{\'a}k, Peter
                and Retkowski, Fabian
                and Savoldi, Beatrice
                and Sikasote, Claytone
                and Sperber, Matthias
                and St{\"u}ker, Sebastian
                and Sudoh, Katsuhito
                and Tahon, Marie
                and Turchi, Marco
                and Waibel, Alex
                and Wilken, Patrick
                and Zevallos, Rodolfo
                and Zouhar, Vil{\'e}m
                and Z{\"u}fle, Maike
                },
    booktitle = {Proc. of IWSLT},
    year      = {2026},
}

\appendix

\section{Appendix}
\label{sec:appendix}

\begin{table*}[t]
  \centering
  \small
  \begin{threeparttable}[b]
    \begin{tabular}{>{\centering\arraybackslash}p{4em}ccccc}
    \hline
    \textbf{Dataset} & \textbf{Task} & \textbf{Language} & \textbf{Samples} & \textbf{Total} & \textbf{Mean}\\
    \hline
    \multirow{8}{*}{CoVoST}
    & ASR\tnote{O}          & en-en & 289,412 & 429.61h & 5.34s \\
    & ST\tnote{O}           & en-de & 289,412 & 429.61h & 5.34s \\
    & ST\tnote{S}           & en-it & 222,604 & 333.08h & 5.39s \\
    & ST\tnote{O}           & en-zh & 289,412 & 429.61h & 5.34s \\
    \cdashline{2-6}
    & SQA\tnote{S}          & en    & 303,809 & 450.90h & 5.34s \\
    & SQA\tnote{S}          & de    & 300,165 & 444.59h & 5.33s \\
    & SQA\tnote{S}          & it    & 231,479 & 345.74h & 5.38s \\
    & SQA\tnote{S}          & zh    & 303,735 & 450.83h & 5.34s \\
    \hline
    \multirow{3}{*}{EuroParlST}
    & ASR\tnote{O}          & en-en & 62,180 & 151.56h & 8.61s \\
    & ST\tnote{O}           & en-de & 32,628 &  77.17h & 8.51s \\
    & ST\tnote{O}           & en-it & 29,552 &  74.39h & 9.06s \\
    \hline
    \multirow{3}{*}{GigaST}
    & ASR\tnote{O}          & en-en & 909,728 & 997.96h & 3.95s \\
    & ST\tnote{O}           & en-de & 570,183 & 650.08h & 4.10s \\
    & ST\tnote{O}           & en-zh & 444,506 & 489.43h & 3.96s \\
    \hline
    \multirow{8}{*}{LibriSQA}
    & ASR\tnote{O}          & en-en & 109,214 & 381.82h & 12.59s \\
    & ST\tnote{S}           & en-de &  86,816 & 294.69h & 12.22s \\
    & ST\tnote{S}           & en-it &  85,072 & 287.66h & 12.17s \\
    & ST\tnote{S}           & en-zh & 107,982 & 376.87h & 12.56s \\
    \cdashline{2-6}
    & SQA\tnote{O}          & en    & 109,214 & 381.82h & 12.59s \\
    & SQA\tnote{S}          & de    &  86,816 & 294.69h & 12.22s \\
    & SQA\tnote{S}          & it    &  85,072 & 287.66h & 12.17s \\
    & SQA\tnote{S}          & zh    & 107,982 & 376.87h & 12.56s \\
    \hline
    Total & - & - & 5,056,973 & 8,436.64h & 6.01s\\
    \hline
    \end{tabular}
    \begin{tablenotes}[flushleft]
    \tiny
    \begingroup
    \setlength{\multicolsep}{0pt}
    \setlength{\premulticols}{0pt}
    \setlength{\postmulticols}{0pt}
    \setlength{\columnsep}{3em}
    \raggedcolumns
    \begin{multicols}{2}
    \item[O] Original data provided by the dataset.
    \item[S] Synthetic data generated by the models.
    \end{multicols}
    \endgroup
    \end{tablenotes}
  \end{threeparttable}
  \caption{Statistics of the training corpora for the IWSLT short track under
    constrained settings.}
  \label{tab:short-corpora}
\end{table*}

\begin{table*}[t]
  \centering
  \small
  \begin{threeparttable}[b]
    \begin{tabular}{>{\centering\arraybackslash}p{4em}ccccc}
    \hline
    \textbf{Dataset} & \textbf{Task} & \textbf{Language} & \textbf{Samples} & \textbf{Total} & \textbf{Mean}\\
    \hline
    \multirow{8}{*}{LibriSQA}
    & Short                 &     - & 778,168 & 2,682.08h &  12.41s \\
    \cdashline{2-6}
    & ASR\tnote{O}          & en-en &   1,502 &   207.60h & 497.58s \\
    & ST\tnote{S}           & en-de &   1,786 &   209.18h & 421.64s \\
    & ST\tnote{S}           & en-it &   1,808 &   207.73h & 413.63s \\
    & ST\tnote{S}           & en-zh &   1,515 &   207.47h & 493.00s \\
    \cdashline{2-6}
    & SQA\tnote{O}          &    en &   1,502 &   207.60h & 497.58s \\
    & SQA\tnote{S}          &    de &   1,786 &   209.18h & 421.64s \\
    & SQA\tnote{S}          &    it &   1,808 &   207.73h & 413.63s \\
    & SQA\tnote{S}          &    zh &   1,515 &   207.47h & 493.00s \\
    \hline
    \multirow{3}{*}{Nutshell}
    & ASR\tnote{O}          & en-en &   3,139 &   501.94h & 575.66s \\
    & ST\tnote{S}           & en-de &   2,363 &   374.70h & 570.85s \\
    & ST\tnote{S}           & en-it &   2,675 &   426.57h & 574.07s \\
    & ST\tnote{S}           & en-zh &   2,371 &   379.01h & 575.46s \\
    \hline
    \multirow{3}{*}{YTSeg}
    & ASR\tnote{O}          & en-en &   6,243 &   694.63h & 400.56s \\
    & ST\tnote{S}           & en-de &   6,091 &   679.68h & 401.72s \\
    & ST\tnote{S}           & en-it &   5,921 &   662.75h & 402.95s \\
    & ST\tnote{S}           & en-zh &   5,063 &   570.96h & 405.98s \\
    \hline
    Total & - & - & 825,256 & 9,496.36h & 41.43s \\
    \hline
    \end{tabular}
    \begin{tablenotes}[flushleft]
    \tiny
    \begingroup
    \setlength{\multicolsep}{0pt}
    \setlength{\premulticols}{0pt}
    \setlength{\postmulticols}{0pt}
    \setlength{\columnsep}{3em}
    \raggedcolumns
    \begin{multicols}{2}
    \item[O] Original data provided by the dataset.
    \item[S] Synthetic data generated by the models.
    \end{multicols}
    \endgroup
    \end{tablenotes}
  \end{threeparttable}
  \caption{Statistics of the training corpora for the IWSLT long track under
    constrained settings.}
  \label{tab:long-corpora}
\end{table*}


\begin{table*}
  \centering
  \small
  \begin{tabular}{ccccc}
    \hline
    \multirow{2}{*}{\textbf{Submission}} & \multicolumn{4}{c}{\textbf{EN-EN}} \\
    & \textbf{ASR.WER} & \textbf{SQA.BERTScore} & \textbf{QE.Accuracy} & \textbf{QE.Format.Accuracy} \\
    \hline
    Primary     & 0.123 & 0.507 & - & - \\
    Contrastive & 0.145 & 0.505 & - & - \\
    \hline
    \multirow{2}{*}{\textbf{Submission}} & \multicolumn{4}{c}{\textbf{EN-DE}} \\
    & \textbf{ST.COMET} & \textbf{SQA.BERTScore} & \textbf{QE.Accuracy} & \textbf{QE.Format.Accuracy} \\
    \hline
    Primary     & 0.762 & 0.477 & 0.762 & 0.974 \\
    Contrastive & 0.739 & 0.505 & 0.501 & 0.584 \\
    \hline
    \multirow{2}{*}{\textbf{Submission}} & \multicolumn{4}{c}{\textbf{EN-IT}} \\
    & \textbf{ST.COMET} & \textbf{SQA.BERTScore} & \textbf{QE.Accuracy} & \textbf{QE.Format.Accuracy} \\
    \hline
    Primary     & 0.742 & 0.524 & - & - \\
    Contrastive & 0.733 & 0.527 & - & - \\
    \hline
    \multirow{2}{*}{\textbf{Submission}} & \multicolumn{4}{c}{\textbf{EN-ZH}} \\
    & \textbf{ST.COMET} & \textbf{SQA.BERTScore} & \textbf{QE.Accuracy} & \textbf{QE.Format.Accuracy} \\
    \hline
    Primary     & 0.777 & 0.520 & 0.915 & 0.961 \\
    Contrastive & 0.734 & 0.482 & 0.658 & 0.819 \\
    \hline
  \end{tabular}
  \caption{Performance of the primary and contrastive submissions on the IWSLT 2026 short-form tasks.}
  \label{tab:iwslt-short}
\end{table*}

\begin{table*}[t]
  \centering
  \scriptsize
  \resizebox{\linewidth}{!}{
  \begin{threeparttable}
    \begin{tabular}{ccccccccccc}
    \hline
    \multirow{3}{*}{\textbf{Submission}} & \multicolumn{10}{c}{\textbf{EN-EN}} \\
    & \textbf{ASR} & \textbf{SQA} & \textbf{SSUM} & \textbf{QE} & \textbf{QE.Format} & \textbf{ACHAP} & \textbf{ACHAP} & \textbf{ACHAP.GC} & \textbf{ACHAP.TM} & \textbf{ACHAP.TM} \\
    & \textbf{WER} & \textbf{BERTScore} & \textbf{BERTScore} & \textbf{Accuracy} & \textbf{Accuracy} & \textbf{WER} & \textbf{CollarF1} & \textbf{BERTScore} & \textbf{BERTScore} & \textbf{Matched} \\
    \hline
    Fixed  & 0.196 & 0.390 & 0.152 & - & - & 0.359 & 0.000 (0.183)\tnote{\dag} & 0.804 (0.823)\tnote{\dag} & - & - \\
    CRDNN  & 0.175 & 0.377 & 0.160 & - & - & 0.200 & 0.000 (0.271)\tnote{\dag} & 0.803 (0.842)\tnote{\dag} & - & - \\
    Hybrid & 0.126 & 0.377 & 0.156 & - & - & 0.200 & 0.000 (0.271)\tnote{\dag} & 0.803 (0.842)\tnote{\dag} & - & - \\
    \hline
    \multirow{3}{*}{\textbf{Submission}} & \multicolumn{10}{c}{\textbf{EN-DE}} \\
    & \textbf{ST} & \textbf{SQA} & \textbf{SSUM} & \textbf{QE} & \textbf{QE.Format} & \textbf{ACHAP} & \textbf{ACHAP} & \textbf{ACHAP.GC} & \textbf{ACHAP.TM} & \textbf{ACHAP.TM} \\
    & \textbf{COMET} & \textbf{BERTScore} & \textbf{BERTScore} & \textbf{Accuracy} & \textbf{Accuracy} & \textbf{COMET} & \textbf{CollarF1} & \textbf{BERTScore} & \textbf{BERTScore} & \textbf{Matched} \\
    \hline
    Fixed  & 0.694 & 0.348 & 0.149 & 0.501 & 0.584 & 0.690 & 0.000 (0.195)\tnote{\dag} & 0.584 (0.618)\tnote{\dag} & 0.793 & 0.127 \\
    CRDNN  & 0.722 & 0.351 & 0.153 & 0.501 & 0.584 & 0.635 & 0.000 (0.136)\tnote{\dag} & 0.616 (0.640)\tnote{\dag} & 0.789 & 0.135 \\
    Hybrid & 0.723 & 0.350 & 0.153 & 0.501 & 0.584 & 0.647 & 0.000 (0.165)\tnote{\dag} & 0.616 (0.650)\tnote{\dag} & 0.789 & 0.135 \\
    \hline
    \multirow{3}{*}{\textbf{Submission}} & \multicolumn{10}{c}{\textbf{EN-IT}} \\
    & \textbf{ST} & \textbf{SQA} & \textbf{SSUM} & \textbf{QE} & \textbf{QE.Format} & \textbf{ACHAP} & \textbf{ACHAP} & \textbf{ACHAP.GC} & \textbf{ACHAP.TM} & \textbf{ACHAP.TM} \\
    & \textbf{COMET} & \textbf{BERTScore} & \textbf{BERTScore} & \textbf{Accuracy} & \textbf{Accuracy} & \textbf{COMET} & \textbf{CollarF1} & \textbf{BERTScore} & \textbf{BERTScore} & \textbf{Matched} \\
    \hline
    Fixed  & 0.707 & 0.385 & 0.185 & - & - & 0.723 & 0.000 (0.335)\tnote{\dag} & 0.532 (0.597)\tnote{\dag} & 0.773 & 0.123 \\
    CRDNN  & 0.702 & 0.376 & 0.171 & - & - & 0.719 & 0.000 (0.348)\tnote{\dag }& 0.586 (0.660)\tnote{\dag} & 0.790 & 0.135 \\
    Hybrid & 0.695 & 0.373 & 0.174 & - & - & 0.735 & 0.000 (0.315)\tnote{\dag} & 0.528 (0.665)\tnote{\dag} & 0.767 & 0.123 \\
    \hline
    \multirow{3}{*}{\textbf{Submission}} & \multicolumn{10}{c}{\textbf{EN-ZH}} \\
    & \textbf{ST} & \textbf{SQA} & \textbf{SSUM} & \textbf{QE} & \textbf{QE.Format} & \textbf{ACHAP} & \textbf{ACHAP} & \textbf{ACHAP.GC} & \textbf{ACHAP.TM} & \textbf{ACHAP.TM} \\
    & \textbf{COMET} & \textbf{BERTScore} & \textbf{BERTScore} & \textbf{Accuracy} & \textbf{Accuracy} & \textbf{COMET} & \textbf{CollarF1} & \textbf{BERTScore} & \textbf{BERTScore} & \textbf{Matched} \\
    \hline
    Fixed  & 0.655 & 0.380 & 0.319 & 0.658 & 0.819 & 0.681 & 0.000 (0.073)\tnote{\dag} & 0.496 (0.518)\tnote{\dag} & 0.757 & 0.135 \\
    CRDNN  & 0.699 & 0.374 & 0.325 & 0.658 & 0.819 & 0.698 & 0.000 (0.071)\tnote{\dag} & 0.496 (0.522)\tnote{\dag} & 0.757 & 0.135 \\
    Hybrid & 0.685 & 0.373 & 0.325 & 0.658 & 0.819 & 0.683 & 0.000 (0.021)\tnote{\dag} & 0.496 (0.506)\tnote{\dag} & 0.757 & 0.135 \\
    \hline
  \end{tabular}
  \begin{tablenotes}
  \tiny
  \item[\dag] Values in parentheses are obtained under a relaxed Markdown-format evaluation.
  \end{tablenotes}
  \end{threeparttable}
  }
  \caption{Performance of the primary and contrastive submissions on the IWSLT 2026 long-form tasks.}
  \label{tab:iwslt-long}
\end{table*}

\begin{table*}
  \centering
  \small
  \begin{tabular}{cccccccccc}
    \hline
    \multirow{2}{*}{\textbf{CRDNN}}
    & \multirow{2}{*}{\textbf{Language}}
    & \multicolumn{2}{c}{\textbf{ASR}}
    & \multicolumn{2}{c}{\textbf{ST}}
    & \multicolumn{2}{c}{\textbf{SQA}}
    & \multicolumn{2}{c}{\textbf{SUM}} \\
    & & \textbf{WER} & \textbf{HAL}
    & \textbf{COMET} & \textbf{HAL}
    & \textbf{BERTScore} & \textbf{HAL}
    & \textbf{BERTScore} & \textbf{HAL} \\
    \hline
    \multirow{4}{*}{-Hallucination}
    & en-en & 0.1580 & 2/21 &      - &    - & 0.3791 & 0/220 & 0.1860 & 16/100 \\
    & en-de &      - &    - & 0.7248 & 2/21 & 0.3748 & 3/220 & 0.1861 & 29/100 \\
    & en-it &      - &    - & 0.7707 & 2/21 & 0.3311 & 3/220 & 0.2351 & 38/100 \\
    & en-zh &      - &    - & 0.7630 & 5/21 & 0.4009 & 3/220 & 0.3632 & 16/100 \\
    \hline
    \multirow{4}{*}{+Hallucination}
    & en-en & 1.0161 & 2/21 &      - &    - & 0.3791 & 0/220 & 0.1643 & 16/100 \\
    & en-de &      - &    - & 0.6936 & 2/21 & 0.3684 & 3/220 & 0.1476 & 29/100 \\
    & en-it &      - &    - & 0.7480 & 2/21 & 0.3241 & 3/220 & 0.1712 & 38/100 \\
    & en-zh &      - &    - & 0.6841 & 5/21 & 0.3948 & 3/220 & 0.3473 & 16/100 \\
    \hline
    $|\mathrm{diff}|$ & - & 0.8581 & - & 0.1328 & - & 0.0195 & - & 0.1400 & - \\
    \hline
  \end{tabular}
  \caption{Performance of CRDNN segmentation +/-hallucination with $(0.5, 0.25,
    0.5)$ on the MCIF long-form tasks.}
  \label{tab:mcif-hallucination}
\end{table*}

\begin{table*}
  \centering
  \scriptsize
  \begin{tabular}{ccccccccccc}
    \hline
    \multirow{2}{*}{\textbf{Window}}
    & \multirow{2}{*}{\textbf{Language}}
    & \multicolumn{2}{c}{\textbf{ASR}}
    & \multicolumn{2}{c}{\textbf{ST}}
    & \multicolumn{2}{c}{\textbf{SQA}}
    & \multicolumn{2}{c}{\textbf{SUM}} \\
    & & \textbf{Accuracy} & \textbf{HAL}
    & \textbf{COMET} & \textbf{HAL}
    & \textbf{BERTScore} & \textbf{HAL}
    & \textbf{BERTScore} & \textbf{HAL}
    & \textbf{HIFS} \\
    \hline
    \multirow{5}{*}{15s}
    & en-en & 0.8737 & 3/21 &      - &    - & 0.3800 & 1/220 & 0.1929 & 12/100 & - \\
    & en-de &      - &    - & 0.7124 & 3/21 & 0.3756 & 4/220 & 0.1884 & 26/100 & - \\
    & en-it &      - &    - & 0.7667 & 3/21 & 0.3496 & 0/220 & 0.2252 & 27/100 & - \\
    & en-zh &      - &    - & 0.7595 & 4/21 & 0.3927 & 3/220 & 0.3578 &  9/100 & - \\
    \cdashline{2-11}
    & score & 0.7489 & 3/21 & 0.6275 & 10/63 & \textbf{0.3710} & \textbf{8/440} & 0.1998 & 74/400 & 1.9472 \\
    \hline
    \multirow{5}{*}{30s}
    & en-en & 0.8582 & 0/21 &      - &    - & 0.3755 & 1/220 & 0.1850 &  7/100 & - \\
    & en-de &      - &    - & 0.7193 & 3/21 & 0.3699 & 0/220 & 0.2008 & 25/100 & - \\
    & en-it &      - &    - & 0.7546 & 1/21 & 0.3356 & 4/220 & 0.2203 & 31/100 & - \\
    & en-zh &      - &    - & 0.7365 & 4/21 & 0.3937 & 4/220 & 0.3586 & 10/100 & - \\
    \cdashline{2-11}
    & score & \textbf{0.8582} & \textbf{0/21} & \textbf{0.6438} & \textbf{8/63} & 0.3649 & 9/440 & 0.1993 & 73/400 & \textbf{2.0663} \\
    \hline
    \multirow{5}{*}{45s}
    & en-en & 0.8762 & 2/21 &      - &    - & 0.3795 & 1/220 & 0.1917 & 13/100 & - \\
    & en-de &      - &    - & 0.6636 & 5/21 & 0.3664 & 1/220 & 0.1917 & 18/100 & - \\
    & en-it &      - &    - & 0.7557 & 2/21 & 0.3369 & 3/220 & 0.2402 & 34/100 & - \\
    & en-zh &      - &    - & 0.7608 & 7/21 & 0.3962 & 6/220 & 0.3623 &  5/100 & - \\
    \cdashline{2-11}
    & score & 0.7928 & 2/21 & 0.5655 & 14/63 & 0.3651 & 11/440 & \textbf{0.2067} & \textbf{70/400} & 1.9300 \\
    \hline
    \multirow{5}{*}{60s}
    & en-en & 0.8679 & 2/21 &      - &    - & 0.3698 & 0/220 & 0.1850 &  8/100 & - \\
    & en-de &      - &    - & 0.7067 & 2/21 & 0.3573 & 2/220 & 0.1994 & 27/100 & - \\
    & en-it &      - &    - & 0.7611 & 4/21 & 0.3266 & 1/220 & 0.2377 & 38/100 & - \\
    & en-zh &      - &    - & 0.7529 & 5/21 & 0.3956 & 6/220 & 0.3562 &  7/100 & - \\
    \cdashline{2-11}
    & score & 0.7852 & 2/21 & 0.6097 & 11/63 & 0.3584 & 9/440 & 0.1986 & 80/400 & 1.9520 \\
    \hline
  \end{tabular}
  \caption{Performance of different window sizes on the MCIF long-form tasks,
    together with hallucination (HAL) statistics, where the 30-second window
    achieves the best overall performance.}
  \label{tab:mcif-window}
\end{table*}

\begin{table*}
  \centering
  \scriptsize
  \begin{tabular}{ccccccccccc}
    \hline
    \multirow{2}{*}{\textbf{Parameters}}
    & \multirow{2}{*}{\textbf{Language}}
    & \multicolumn{2}{c}{\textbf{ASR}}
    & \multicolumn{2}{c}{\textbf{ST}}
    & \multicolumn{2}{c}{\textbf{SQA}}
    & \multicolumn{2}{c}{\textbf{SUM}} \\
    & & \textbf{Accuracy} & \textbf{HAL}
    & \textbf{COMET} & \textbf{HAL}
    & \textbf{BERTScore} & \textbf{HAL}
    & \textbf{BERTScore} & \textbf{HAL}
    & \textbf{HIFS} \\
    \hline
    \multirow{5}{*}{(0.5, 0.25, 0.5)}
    & en-en & 0.8420 & 2/21 &      - &    - & 0.3791 & 0/220 & 0.1860 & 16/100 & - \\
    & en-de &      - &    - & 0.7248 & 2/21 & 0.3748 & 3/220 & 0.1861 & 29/100 & - \\
    & en-it &      - &    - & 0.7707 & 2/21 & 0.3311 & 3/220 & 0.2351 & 38/100 & - \\
    & en-zh &      - &    - & 0.7630 & 5/21 & 0.4009 & 3/220 & 0.3632 & 16/100 & - \\
    \cdashline{2-11}
    & score & 0.7618 & \textbf{2/21} & \textbf{0.6448} & \textbf{9/63} & 0.3677 & 9/440 & 0.1848 & 99/400 & \textbf{1.9591} \\
    \hline
    \multirow{5}{*}{(0.3, 0.2, 0.3)}
    & en-en & 0.8421 & 2/21 &      - &    - & 0.3755 & 0/220 & 0.1932 & 11/100 & - \\
    & en-de &      - &    - & 0.7035 & 2/21 & 0.3795 & 0/220 & 0.1913 & 21/100 & - \\
    & en-it &      - &    - & 0.7749 & 5/21 & 0.3320 & 2/220 & 0.2205 & 28/100 & - \\
    & en-zh &      - &    - & 0.7696 & 5/21 & 0.4041 & 2/220 & 0.3573 &  2/100 & - \\
    \cdashline{2-11}
    & score & \textbf{0.7619} & \textbf{2/21} & 0.6044 & 12/63 & \textbf{0.3711} & \textbf{4/880} & \textbf{0.2080} & \textbf{62/400} & 1.9454 \\
    \hline
    \multirow{5}{*}{(0.2, 0.1, 0.2)}
    & en-en & 0.8367 & 3/21 &      - &    - & 0.3666 & 1/220 & 0.2025 & 10/100 & - \\
    & en-de &      - &    - & 0.7116 & 3/21 & 0.3683 & 0/220 & 0.2044 & 29/100 & - \\
    & en-it &      - &    - & 0.7578 & 4/21 & 0.3331 & 0/220 & 0.2353 & 38/100 & - \\
    & en-zh &      - &    - & 0.7727 & 5/21 & 0.4002 & 3/220 & 0.3547 &  8/100 & - \\
    \cdashline{2-11}
    & score & 0.7172 & 3/21 & 0.6040 & 11/63 & 0.3653 & 9/440 & 0.1999 & 80/400 & 1.8864 \\
    \hline
  \end{tabular}
  \caption{Performance of different parameters with the CRDNN segmentation on
    the MCIF long-form tasks, together with hallucination (HAL) statistics,
    where the (0.5, 0.25, 0.5) setting achieves the best overall performance.}
  \label{tab:mcif-crdnn}
\end{table*}

\begin{table*}
  \centering
  \scriptsize
  \begin{tabular}{ccccccccccc}
    \hline
    \multirow{2}{*}{\textbf{Parameters}}
    & \multirow{2}{*}{\textbf{Language}}
    & \multicolumn{2}{c}{\textbf{ASR}}
    & \multicolumn{2}{c}{\textbf{ST}}
    & \multicolumn{2}{c}{\textbf{SQA}}
    & \multicolumn{2}{c}{\textbf{SUM}} \\
    & & \textbf{Accuracy} & \textbf{HAL}
    & \textbf{COMET} & \textbf{HAL}
    & \textbf{BERTScore} & \textbf{HAL}
    & \textbf{BERTScore} & \textbf{HAL}
    & \textbf{HIFS} \\
    \hline
    \multirow{5}{*}{(0.5, 0.25, 0.5)}
    & en-en & 0.8430 & 2/21 &      - &    - & 0.3807 & 0/220 & 0.1844 & 14/100 & - \\
    & en-de &      - &    - & 0.7189 & 3/21 & 0.3747 & 3/220 & 0.1874 & 22/100 & - \\
    & en-it &      - &    - & 0.7739 & 1/21 & 0.3332 & 4/220 & 0.2192 & 29/100 & - \\
    & en-zh &      - &    - & 0.7612 & 5/21 & 0.3990 & 3/220 & 0.3620 & 16/100 & - \\
    \cdashline{2-11}
    & score & 0.7627 & 2/21 & \textbf{0.6444} & \textbf{9/63} & 0.3677 & 10/440 & 0.1911 & 81/400 & 1.9660 \\
    \hline
    \multirow{5}{*}{(0.3, 0.2, 0.3)}
    & en-en & 0.8431 & 1/21 &      - &    - & 0.3728 & 0/220 & 0.1975 & 11/100 & - \\
    & en-de &      - &    - & 0.7043 & 1/21 & 0.3747 & 0/220 & 0.2024 & 28/100 & - \\
    & en-it &      - &    - & 0.7761 & 4/21 & 0.3333 & 3/220 & 0.2335 & 38/100 & - \\
    & en-zh &      - &    - & 0.7678 & 4/21 & 0.4038 & 2/220 & 0.3545 &  4/100 & - \\
    \cdashline{2-11}
    & score & \textbf{0.8030} & \textbf{1/21} & 0.6402 & \textbf{9/63} & \textbf{0.3691} & \textbf{5/880} & 0.2016 & 81/400 & \textbf{2.0139} \\
    \hline
    \multirow{5}{*}{(0.2, 0.1, 0.2)}
    & en-en & 0.7692 & 2/21 &      - &    - & 0.3668 & 0/220 & 0.1995 &  7/100 & - \\
    & en-de &      - &    - & 0.6989 & 2/21 & 0.3701 & 3/220 & 0.1944 & 24/100 & - \\
    & en-it &      - &    - & 0.7576 & 4/21 & 0.3346 & 1/220 & 0.2136 & 27/100 & - \\
    & en-zh &      - &    - & 0.7734 & 5/21 & 0.3997 & 2/220 & 0.3529 &  6/100 & - \\
    \cdashline{2-11}
    & score & 0.6959 & 2/21 & 0.6116 & 11/63 & 0.3652 & 6/440 & \textbf{0.2052} & \textbf{64/400} & 1.8781 \\
    \hline
  \end{tabular}
  \caption{Performance of different parameters with the hybrid segmentation on
    the MCIF long-form tasks, together with hallucination (HAL) statistics,
    where the (0.3, 0.2, 0.3) setting achieves the best overall performance.}
  \label{tab:mcif-hybrid}
\end{table*}

\end{document}